\documentclass[runningheads]{llncs}
\usepackage{amsmath,epsfig}
\usepackage{graphicx} 
\usepackage{times}  
\usepackage{makecell}
\usepackage{tabularx}
\usepackage{placeins}
\usepackage{cite}
\usepackage{amssymb}
\usepackage{newfloat}
\usepackage{marvosym}
\usepackage{ifsym}
\begin{document}
\title{Towards Balanced RGB-TSDF Fusion for Consistent Semantic Scene Completion by 3D RGB Feature Completion and a Classwise Entropy Loss Function}
%
%
\author{Laiyan Ding\inst{1}\orcidID{0009-0006-3093-5335} \and
Panwen Hu\inst{1}\orcidID{0000-0001-6183-6598} \and \\
Jie Li\inst{2}\orcidID{0000-0001-6254-1724} \and
Rui Huang \inst{1{\textrm{\Letter}}}\orcidID{0000-0002-7950-1662}}
\authorrunning{L. Ding et al.}
\titlerunning{Towards Balanced RGB-TSDF Fusion for Consistent Semantic Scene Completion}
%
	\institute{ School of Science and Engineering, The Chinese University of Hong Kong (Shenzhen), Shenzhen, Guangdong, China\\
	\email{\{laiyanding,panwenhu\}@link.cuhk.edu.cn}, \email{ruihuang@cuhk.edu.cn}   \and
	School of Artificial Intelligence, Shenzhen Polytechnic University, Shenzhen, Guangdong, China \\
	\email{jieli1@szpt.edu.cn}}
\maketitle              
\begin{abstract}
	Semantic Scene Completion (SSC) aims to jointly infer semantics and occupancies of 3D scenes. Truncated Signed Distance Function (TSDF), a 3D encoding of depth, has been a common input for SSC. Furthermore, RGB-TSDF fusion, seems promising since these two modalities provide color and geometry information, respectively. Nevertheless, RGB-TSDF fusion has been considered nontrivial and commonly-used naive addition will result in inconsistent results. We argue that the inconsistency comes from the sparsity of RGB features upon projecting into 3D space, while TSDF features are dense, leading to imbalanced feature maps when summed up. To address this RGB-TSDF distribution difference, we propose a two-stage network with a 3D RGB feature completion module that completes RGB features with meaningful values for occluded areas. Moreover, we propose an effective classwise entropy loss function to punish inconsistency. Extensive experiments on public datasets verify that our method achieves state-of-the-art performance among methods that do not adopt extra data. 
	
	\keywords{Semantic Scene Completion  \and RGB-TSDF fusion \and Entropy-based loss function.}
\end{abstract}

\section{Introduction}
Semantic scene completion is the task that reconstructs the entire scene and infers the voxel-wise semantics, including both visible surfaces and occluded areas, given a single depth image or a pair of RGB and depth images \cite{song2017semantic,garbade2019two}. This ability to infer 3D semantics and geometry from a single 2D observation can benefit diverse computer vision applications, e.g., indoor navigation of robots and autonomous driving. 

In the task of 3D semantic scene completion, the pioneering work SSCNet \cite{song2017semantic} takes Truncated Signed Distance Function (TSDF), an encoding of depth in 3D space, as the sole input. Later works \cite{garbade2019two, guedes2018semantic} propose naive ways to incorporate RGB information into the depth-only network and verify the benefit of including RGB as inputs. RGB provides rich texture and color information and depth describes the local geometry or shape information. These two modalities encode different but complementary information. Afterward, most existing methods take both RGB and depth as inputs and adopt various fusion methodologies \cite{li2019rgbd, wang2022ffnet}. 

\begin{figure}
	\includegraphics[width=1.0\textwidth]{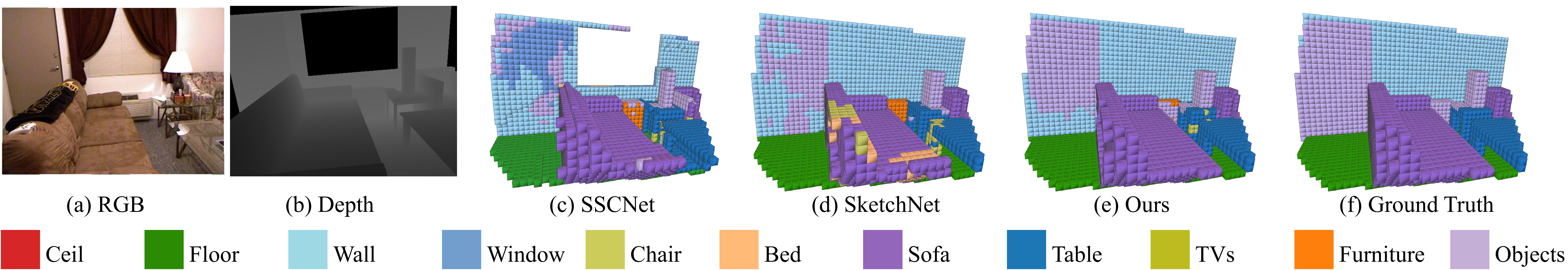}
	\caption{Visualization of semantic scene completion results on NYUCAD dataset. From left to right: (a) RGB input, (b) depth map, (c) results of SSCNet \cite{song2017semantic}, (d) results of SketchNet \cite{chen20203d}, (e) results of our proposed method, (f) ground truth. Our method can achieve better instance consistency on \textit{sofa} and \textit{wall}, which are in occluded areas, compared with SSCNet \cite{song2017semantic} and SketchNet~\cite{chen20203d}. Best viewed in color and zoomed in.}
	\label{fig:firsts}
\end{figure}

\begin{figure}
	\centering
	\includegraphics[width=\textwidth]{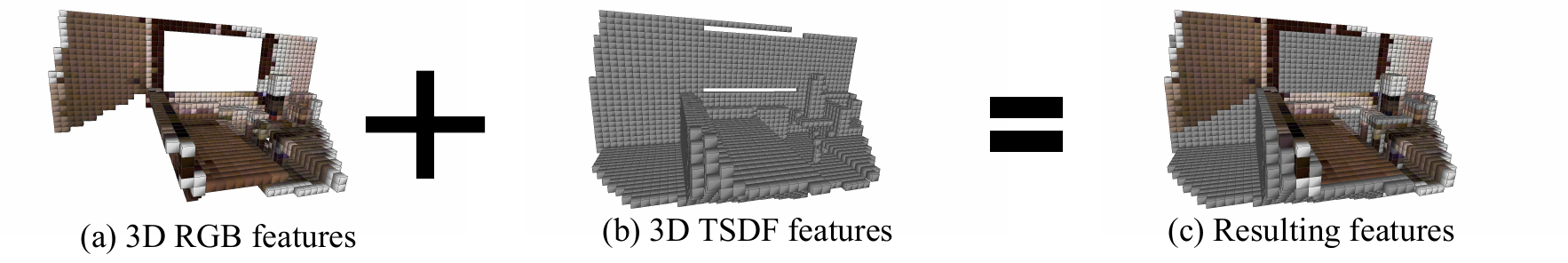}
	\caption{Addition of 3D RGB and TSDF features. We visualize RGB and TSDF in (a) and (b), respectively, for better illustration. In the resulting features (c), we visualize RGB on the visible surfaces and TSDF in occluded areas.}
	\label{fig:difference}
\end{figure}

However, previous works mostly consider fusion in 2D space, ignoring that TSDF can provide a rich supervision signal \cite{roldao20223d}. Nevertheless, RGB-TSDF fusion is nontrivial. EdgeNet \cite{dourado2019edgenet} argues that we cannot encode RGB in 3D in a TSDF-like way since RGB is not binary. Consequently, EdgeNet \cite{dourado2019edgenet} projects 2D image edges into 3D space and applies TSDF-encoding to it making it dense. Yet, RGB information will be greatly reduced since only edge information is preserved. SketchNet \cite{chen20203d} uses TSDF to predict a 3D sketch, a geometry prior, and adds it to projected RGB features. Nevertheless, the sketch locates on semantic edges, while RGB features upon projection only exist on visible surfaces. This distribution difference will still lead to inconsistent results shown in Figure \ref{fig:firsts} (d). Similar to SketchNet \cite{chen20203d}, simply adding the dense TSDF and sparse RGB features will also result in imbalanced feature maps shown in Figure \ref{fig:difference}. TSDF features will dominate the occluded areas leading to the failure of RGB-TSDF fusion.


To tackle the aforementioned distribution difference, we propose a two-stage network that takes RGB and TSDF as inputs. The 3D RGB feature completion stage will produce useful TSDF and completed RGB features with a 3D RGB Feature Completion Module (FCM), which assigns meaningful features for occluded areas based on visible surfaces. Then the refined semantic scene completion stage will produce refined results using the features from the first stage as inputs. To further reduce inconsistency, we propose an effective Classwise Entropy Loss (CEL) which calculates the entropy of the mean probability vector for all classes, as additional supervisions aiming at punishing inconsistency throughout classes. Furthermore, the proposed method achieves state-of-the-art performance on NYUCAD \cite{firman2016structured} dataset among methods not using extra data or iterative learning strategy.

\section{Related Work}

\subsection{Semantic Scene Completion}

SSCNet \cite{song2017semantic} is the first work to introduce the task of Semantic Scene Completion (SSC). They use TSDF as the sole input and infer the complete scene structure and voxel-wise semantics. Though TSDF/Depth-only methods can achieve impressive results \cite{zhang2019cascaded,chen2020real, tang2022not}, it is evident that RGB can help identify objects with color and texture information. Some of later works develop RGB-Depth fusion methods in 2D \cite{wang2022ffnet} or 3D~\cite{li2019rgbd,li2023front}. Nevertheless, they ignore the benefits that TSDF brings, i.e., its dense 3D representation and viewpoint invariance \cite{song2017semantic}, which can be suitable for scene representation. Thus, we only consider methods involving RGB-TSDF fusions here.

RGB-TSDF methods conduct the fusion of RGB and TSDF features in 3D space and seem promising since they provide complementary information. Moreover, TSDF is a meaningful signal for the network to learn scene representation \cite{song2017semantic}. EdgeNet \cite{dourado2019edgenet} reasons that the sparsity of RGB leads to the failure of fusion. They propose fusing TSDF-like encoding of 2D edges from the Canny detector and TSDF and achieve promising improvement. SketchNet \cite{chen20203d} proposes embedding TSDF information as 3D sketch, which is resolution-invariant. MFFNet \cite{fu2022semantic} first trains an 3D object detection network and uses the detection results to help refine coarse SSC predictions. However, these methods still suffer from the sparsity of 3D RGB features and inconsistent completion predictions. Unlike previous methods, we transform 3D RGB features from sparse to dense for balanced RGB-TSDF fusions. Furthermore, one possible way to mitigate the sparsity is by depth completion \cite{yan2022rignet,yan2023desnet} to increase the number of points that can be projected.

\subsection{Intra-class Consistency}

Intra-class consistency is a general problem in many computer vision tasks, e.g., image segmentation \cite{yu2018learning, wang2020intra}, clustering \cite{hsu2015neural, karim2021deep}. Wang et al. \cite{wang2020intra} transfer robust intra-class feature representations from teacher network to student network. For neural deep learning-based clustering, Hsu et al. \cite{hsu2015neural} motivate using pairwise constraints to learn clustering in an end-to-end manner and achieve higher purity.  

We propose a classwise entropy loss similar to the intra-class part of constraint in the work of Hsu et al. \cite{hsu2015neural}. Since RGB features are not balanced between visible surfaces and occluded areas, there exists differences predicting semantic labels for them. Therefore, we hope to reduce the variance of predicted probabilities in each object. We impose such constraint by minimizing the entropy of the average probabilities in an object since entropy is a measurement of uncertainty in probabilistic theory \cite{robinson2008entropy}.

\section{Method}
\begin{figure}[t]
	\centering
	\includegraphics[width=0.98\textwidth]{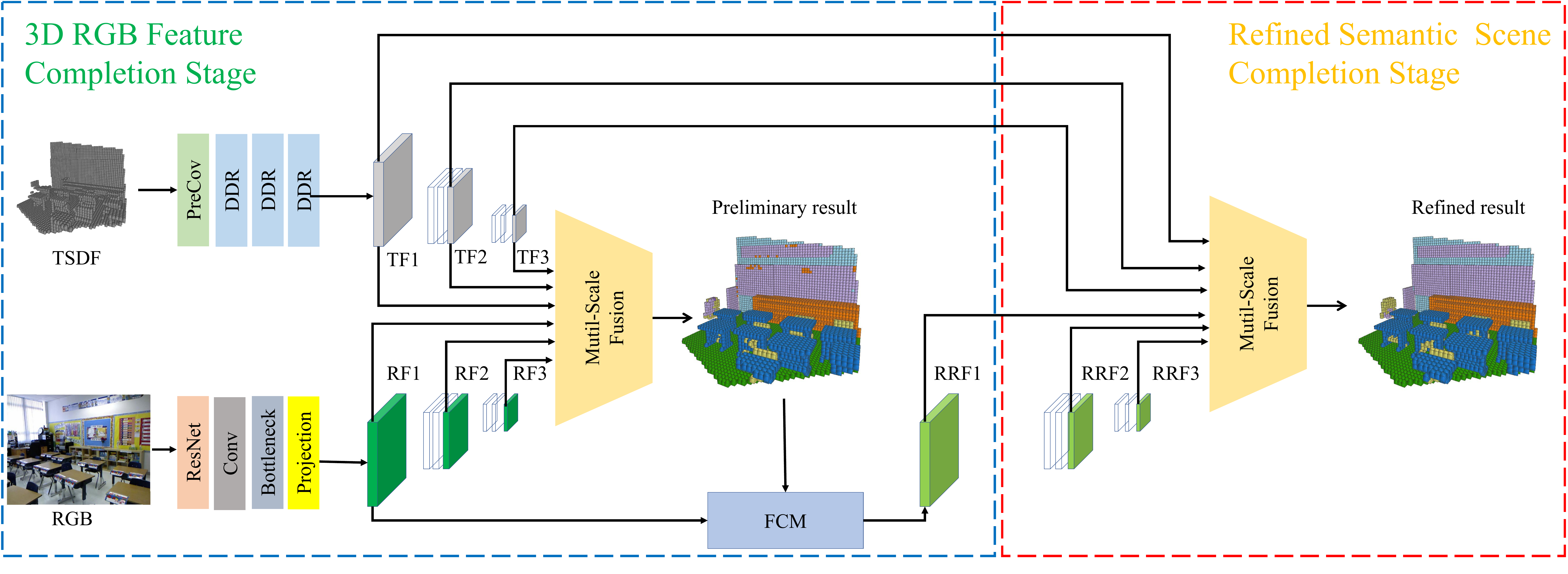}
	\caption{The overview of the proposed network. In 3D RGB feature completion stage, we generate useful TSDF features (TF1, TF2, TF3) and completed 3D RGB features (RRF1) with the proposed 3D RGB Feature Completion Module (FCM). The refined semantic scene completion stage will utilize features from the previous stage to produce the refined result.}
	\label{fig:network}
\end{figure}

In this section, we will first present the structure of our network shown in Figure \ref{fig:network}, then the FCM and the classwise entropy loss for completion and segmentation consistency, and finally, the overall loss function to train our network.

\subsection{Overall Network Architecture}

\subsubsection{3D RGB Feature Completion Stage}
As illustrated in Figure \ref{fig:network},  the 3D RGB feature completion stage takes RGB images and TSDF volumes as inputs. TSDF is computed from a depth image and encodes its information in 3D voxel space where each voxel has a value $d$ indicating the distance to its nearest surface, and the sign of $d$ shows whether the voxel is in free or occluded areas.

In terms of feature extraction, similar to SketchNet \cite{chen20203d}, for TSDF feature extraction, we employ several convolutional layers to transform TSDF into high dimensional features (TF1) for later fusion with RGB features. For RGB feature extraction, a ResNet \cite{he2016deep} is employed as the backbone which is pre-trained on ImageNet \cite{deng2009imagenet} and freezed during training. Next, by utilizing a projection layer similar to SATNet \cite{liu2018see}, we obtain the 3D RGB features in voxel representation which we refer to as RF1.

\begin{figure}[h]
	\centering
	\includegraphics[width=0.6\textwidth]{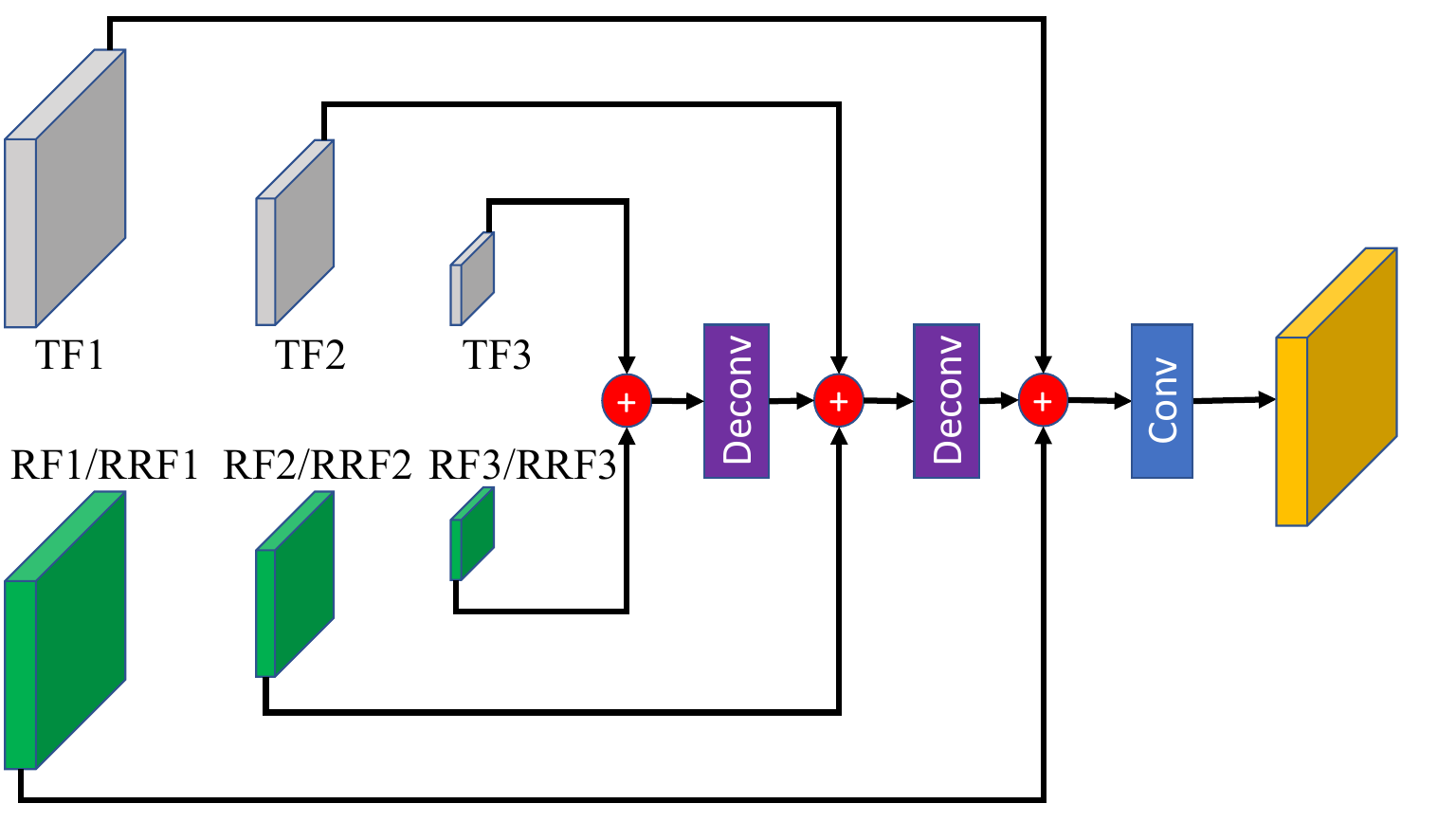}
	\caption{Multi-scale fusion module. This module performs addition and deconvolution in a sequential way.}
	\label{fig:fusion}
\end{figure}

After obtaining features for both modalities in 3D space, i.e., RF1 and TF1, we continue to apply downsampling and convolution to acquire multi-scale feature maps, i.e., RF2, RF3, TF2, TF3. These features are fed into a multi-scale fusion module to generate the preliminary semantic scene completion result. As illustrated in Figure  \ref{fig:fusion}, there is a virtual branch in the middle from adding the two groups of features. This module integrates multi-level and multi-modal features, which have been shown effective for dense predictions \cite{park2017rdfnet}. 

Lastly, the 3D RGB feature completion stage will utilize the preliminary results to transform  sparse RF1 into dense RGB features (RRF1) by feeding them into the proposed 3D RGB feature completion module. 

\subsubsection{Refined Semantic Scene Completion}
The inputs to the refined semantic scene completion stage will be TSDF features from the previous stage, i.e., TF1, TF2, TF3, and the completed RGB features RRF1. After we obtain RRF1, we feed it through successive convolutions similar to stage one. The multi-scale fusion module takes RGB features RRF1, RRF2, RRF3, and TSDF features from stage one and infers the refined result. Reusing TSDF features from the previous stage reduces computation cost and helps refine the result obtained in stage one. Notice we do not reuse RGB features from stage one since convolution layers will learn different weights for sparse and dense inputs.

\subsection{3D RGB Feature Completion Module}
Our motivation is that RGB features throughout an instance should be similar and similar features can lead to similar predictions. Consequently, for each instance, we assign RGB features to occluded areas based on the visible surfaces. Nevertheless, we do not have instance supervision and thus, we conduct the RGB feature completion classwisely. We propose to assign the same feature vector for all the occluded areas using the mean feature vector of visible surfaces for each class. For the sake of simple notations, considering a specific class,
\begin{equation}
	f_j = \frac{1}{|S_v|} \sum_{i \in S_v} {f_i},  \forall{j \in S_o}
\end{equation}
where $f_i$ and $f_j$ are the feature vectors for voxel $i$ and voxel $j$, $S_o$ is the group of occluded voxels, $S_v$ is the group of voxels on the surface. 

\begin{figure}[h]
	\centering
	\includegraphics[width=0.8\textwidth]{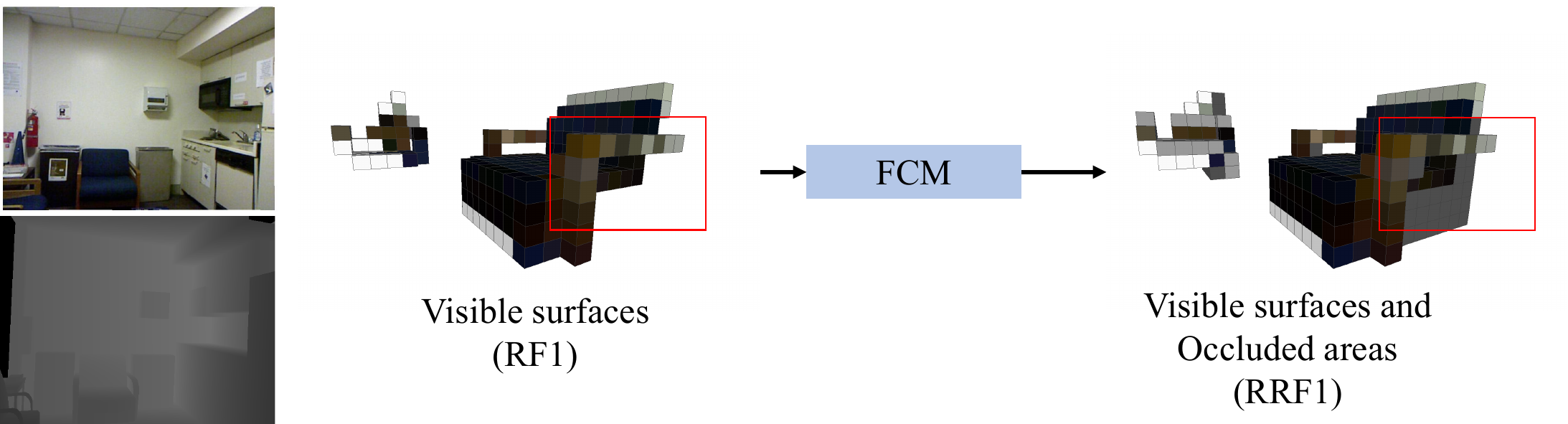}
	\caption{Example of applying FCM on the class \textit{chair}. The 3D RGB feature maps are transformed from sparse to dense.}
	\label{fig:inits}
\end{figure}

The above operation is our 3D RGB FCM. An example on a single class is shown in Figure \ref{fig:inits}, where we detach and transform RF1 to RRF1. By applying FCM, first, the RGB features for the scene excluding seen atmospheres are transformed from sparse to dense since both visible surfaces and occluded areas now contain meaningful features. Second, the RGB features throughout the instances are consistent in semantics and prominence since the features filled in  are the average of features from visible surfaces. In this way, the later fusion with TSDF features will not be dominated by the TSDF. Thus, we can obtain consistent completion results.

Evidently, we cannot inject more information through FCM, since we know nothing about the occluded areas upon projection into 3D space. Nevertheless, this FCM enables easier learning for the RGB branch and, consequently the TSDF branch during the multi-modality fusion.

\subsection{Classwise Entropy Loss}

Imbalanced RGB and TSDF features can lead to inconsistent results, and our FCM can mitigate such imbalance. From another perspective, inconsistent predictions directly come from different predicted probabilities inside instances. Thus, to achieve more consistent predictions for each instance in the scene, we can enforce less variability and randomness in the predicted probabilities for each object. 

Consequently, we impose such constraint by minimizing the entropy of the average probabilities in an object since entropy is a measurement of uncertainty \cite{robinson2008entropy}. Since entropy is a measurement of randomness, reducing entropy reduces the variance, where the entropy is defined as 
\begin{equation}
	Entropy(p) = -\sum_{i} p_{i}log{p_i}
\end{equation}
for any probabilituy vector $p$ and $p_i$ is the i-th element of $p$.

Same as our RGB feature completion method, we conduct the supervision at the class level since we do not have instance-level information. Thus we define our loss as the mean entropy of the mean probability vector for each class, excluding the empty class. Yet we only apply our loss when the class $c$ indeed contains voxels in the ground truth, and the mean probability has the highest probability at position $c$, that is,
\begin{equation}
	\begin{aligned}
		L_E &=\frac{1}{|S|} \sum_{c \in S}Entropy(\frac{1}{N_c} \sum_{i=1}^{N_c} p_{ci}) \\
		&=\frac{1}{|S|} \sum_{c \in S}Entropy(\tilde{p}_c)
	\end{aligned}
\end{equation}
where  $N_c$ is the number of voxels for class $c$ in ground truth, $p_{ci}$ is the predicted probability for voxel $i$ in ground truth class $c$, i.e., $i\in S_c$ and $S_c$ is the voxels that are labeled as class $c$ in ground truth, $\tilde{p}_c$ is the mean probability vector for class $c$ and $S=\{c:N_c>0, argmax(\tilde{p}_c)=c, c>0\}$. In this way, we hope the mean probability to be concentrated and as large as possible at the correct position.

\subsection{Overall Loss function}
Given pairs of RGB-TSDF and 3D ground truth, the total loss function is calculated as:
\begin{equation} 
	\label{eq:4}
	L = L_{ce}^{(1)}+\lambda_1L_{E}^{(1)}+L_{ce}^{(2)}+\lambda_2L_{E}^{(2)}
\end{equation}
where $L_{ce}^{(1)}$,  $L_{E}^{(1)}$, and $\lambda_1$ are cross entropy loss, our classwise entropy loss and weight of the entropy loss for the 3D RGB feature completion stage, respectively. $L_{ce}^{(2)}$,  $L_{E}^{(2)}$, and $\lambda_2$ are defined similarly but for the refined semantic scene completion stage.

\section{Experiments}
\begin{table*}[t]
	\begin{center}
		\caption{Semantic scene completion results on NYUCAD. ED means using extra data, i.e., 2D semantic labels, high resolution ground truth, or 3D instance labels. IL means iterative learning where multiple passes through the network are required. \textbf{Bold} numbers and \underline{underlined} numbers represent the best and the second best scores among similar methods, respectively.}
		\label{tab:table1}
		\resizebox{1.0\textwidth}{!}{
			\begin{tabular}{|l|cc|ccc|cccccccccccc|} 
				\cline{1-18}
				&\multicolumn{2}{|c|}{} & \multicolumn{3}{|c|}{Scene Completion} & \multicolumn{12}{|c|}{Semantic Scene Completion}\\ 
				\cline{1-18}
				
				Methods 			& ED	&	IL	& prec. & recall & IoU & ceil. & floor & wall & win. & chair & bed & sofa & table & tvs & furn & objs & avg.\\ \cline{1-18}
				CCPNet \cite{zhang2019cascaded} &\checkmark  & & 91.3 &  92.6 & 82.4 & 56.2 &  94.6 & 58.7 & 35.1 & 44.8 & 68.6 & 65.3 & 37.6 & 35.5& 53.1 & 35.2 & 53.2 \\
				FFNet \cite{wang2022ffnet} 		&\checkmark & & 94.8 & 90.3 & 85.5 & 62.7 & 94.9 & 67.9 & 35.2 & 52.0 & 74.8 & 69.9 & 47.9 & 27.9& 62.7 & 35.1 & 57.4 \\
				MFFNet \cite{fu2022semantic} & \checkmark & & 88.7 & 92.5 & 84.8 & 54.5 & 94.8 & 63.3 & 29.3 & 50.9 & 73.6 & 70.9 & 56.4 & 31.7 & 61.3 & 42.0 & 57.2 \\ 
				\cline{1-18}
				IMENet \cite{li2021imenet} & \checkmark &\checkmark & 84.8 & 92.3 &79.1 & -& &- &- &- & -& -& -& -& -& -&47.5 \\
				SISNet \cite{cai2021semantic}	& \checkmark &\checkmark & 94.2 & 91.3 & 86.5 & 63.4 & 94.4 & 67.2 & 52.4 & 59.2 & 77.9 & 71.1 & 51.8 & 46.2& 65.8 & 48.8 & 63.5 \\
				\cline{1-18}
				SSCNet \cite{song2017semantic}  &  & & 75.4 &  \textbf{96.3} & 73.2 & 32.5 & 92.6 & 40.2 & 8.9 & 33.9 & 57.0 & 59.5 & 28.3 & 8.1 & 44.8 & 25.1 & 40.0\\
				
				DDRNet \cite{li2019rgbd}		&  & & 88.7 & 88.5 & 79.4 & 54.1 & 91.5 & 56.4 & 14.9 & 37.0 & 55.7 & 51.0 & 28.8 & 9.2 & 44.1 & 27.8 & 42.8\\

				SketchNet \cite{chen20203d}	&  & 	& 90.6 &  \underline{92.2} & \underline{84.2} & 59.7 & \textbf{94.3} & 64.3 & \underline{32.6} & 51.7 & 72.0 & 68.7 & \underline{45.9} & \underline{19.0}& \underline{60.5} & \underline{38.5} & 55.2 \\
				
				PVANet \cite{tang2022not} & & &         \textbf{95.1} & 90.3 & \textbf{86.3} &  \textbf{71.5} & \underline{94.1} & \underline{66.6} & 23.7 & \underline{60.0} & \textbf{78.5} & \underline{72.2} & 45.3 & 16.7& 60.1 & 36.9 & \underline{56.9}
				\\
				Ours							&  & & \underline{94.5} & 87.5 & 83.3 & \underline{64.7} & \textbf{94.3} & \textbf{68.0} & \textbf{35.3} & \textbf{62.7} & \underline{76.9} & \textbf{73.6} & \textbf{49.4} & \textbf{20.9}& \textbf{61.7} & \textbf{41.7} & \textbf{59.0} \\
				\cline{1-18}
		\end{tabular}}
	\end{center}
\end{table*}

\subsection{Implementation Details}
We follow SATNet \cite{liu2018see} to generate TSDF from depth images and downsample the high-resolution 3D ground truth. The overall loss function used is defined as Eq.\ref{eq:4} without extra data, e.g., dense 2D semantic labels and 3D instance information, that originally were not provided in semantic scene completion. Our network is trained with 2 GeForce GTX 2080 Ti GPUs and a batch size of 4 using Pytorch framework. We follow SketchNet \cite{chen20203d} using mini-batch SGD with momentum 0.9 and weight decay 0.0005. The network is trained for 300 epochs. We use a poly learning rate policy and learning rate is updated by $(1-\frac{iteration}{max\_iteration})^{0.9}$. We report the highest mean intersection over union (mIoU) on the validation set among models  saved every 10 epochs.

\subsection{Datasets and Metrics}

Following SketchNet \cite{chen20203d}, we evaluate our method on NYUCAD dataset. NYUCAD is a scene dataset consisting of 1449 indoor scenes. Each sample is a pair of RGB and depth images, and low-resolution ground truth is obtained following SATNet \cite{liu2018see}. NYUCAD provides
depth images projected from 3D annotations to reduce
misalignment and missing depth. There are 795
training samples and 654 test samples.
Following SSCNet \cite{song2017semantic}, we validate our
methods on two tasks, scene completion (SC) and semantic
scene completion (SSC). For SC, we evaluate the voxel-wise predictions, i.e., empty or non-empty on occluded areas. For SSC, we evaluate the
intersection over union (IoU) for each class on both
visible surfaces and occluded areas in the view frustum.

\subsection{Comparisons with State-of-the-art Methods}

\subsubsection{Quantitative Comparison}
We compare our method with other state-of-the-art methods. Table \ref{tab:table1} shows the results on NYUCAD dataset. Our method achieves the best among methods without using extra data for supervision, e.g., 2D semantic labels or 3D instance information. Compared with the recent FFNet \cite{wang2022ffnet} and PVANet \cite{tang2022not}, we obtain an increase of 1.6\%  and 2.1\% on SSC mIoU metric, respectively. Furthermore, we obtain supreme performance on classes that are difficult to maintain consistency during completion, i.e., \textit{chair}, \textit{sofa}, and \textit{wall}. Performance comparisons and analyses on NYU \cite{silberman2012indoor} dataset are in the supplementary material.

\begin{figure}[t]
	
	\includegraphics[width=0.99\textwidth]{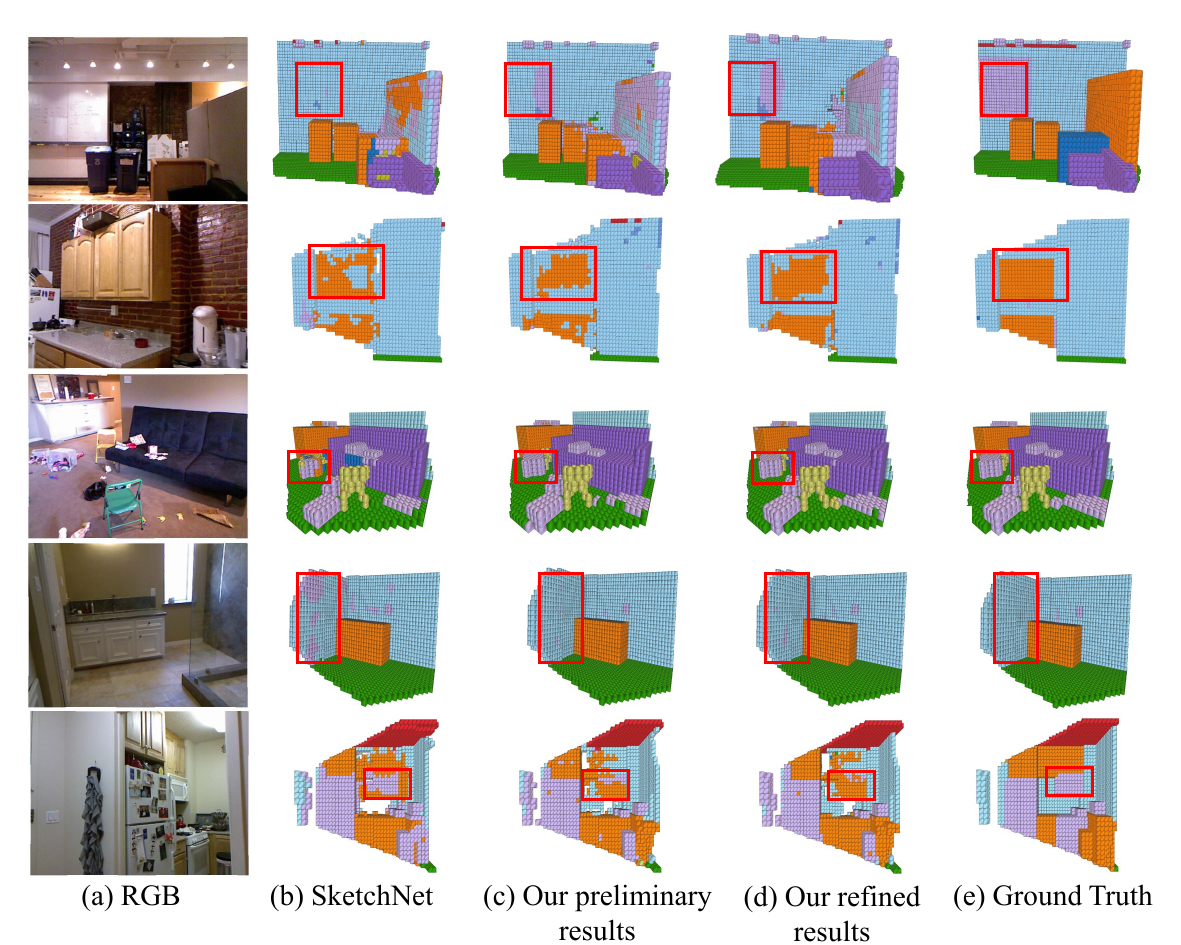}
	\centering
	\caption{Visualization results on NYUCAD dataset. Both FCM and the CEL can help produce consistent results. The second row is the rear view. The last row is a failure case.}
	\label{fig:qualitatives}
\end{figure}

\subsubsection{Qualitative Comparison}
We provide some visualization results on NYUCAD dataset in Figure \ref{fig:qualitatives}. First, we can observe that our preliminary result has been better and more consistent compared with SketchNet \cite{chen20203d}. For example, in the first row, our preliminary result can recover the whiteboard more, and in the third row, we identify the small group of objects as the same class while SkecthNet will predict heterogeneous results. Second, our refined result is superior to the preliminary result. Take the second row, for example, where the predictions are presented in the rear view. Refined results predict more of the cabinet to be \textit{furniture} instead of \textit{wall} since with the proposed FCM, it is easier to predict more coherent results even in occluded areas. Furthermore, with our proposed classwise entropy loss function, the results in both stages are consistent throughout classes, for instance, the fourth row in Figure \ref{fig:qualitatives}.

\begin{figure}[t]
	\centering
	\includegraphics[width=0.9\textwidth]{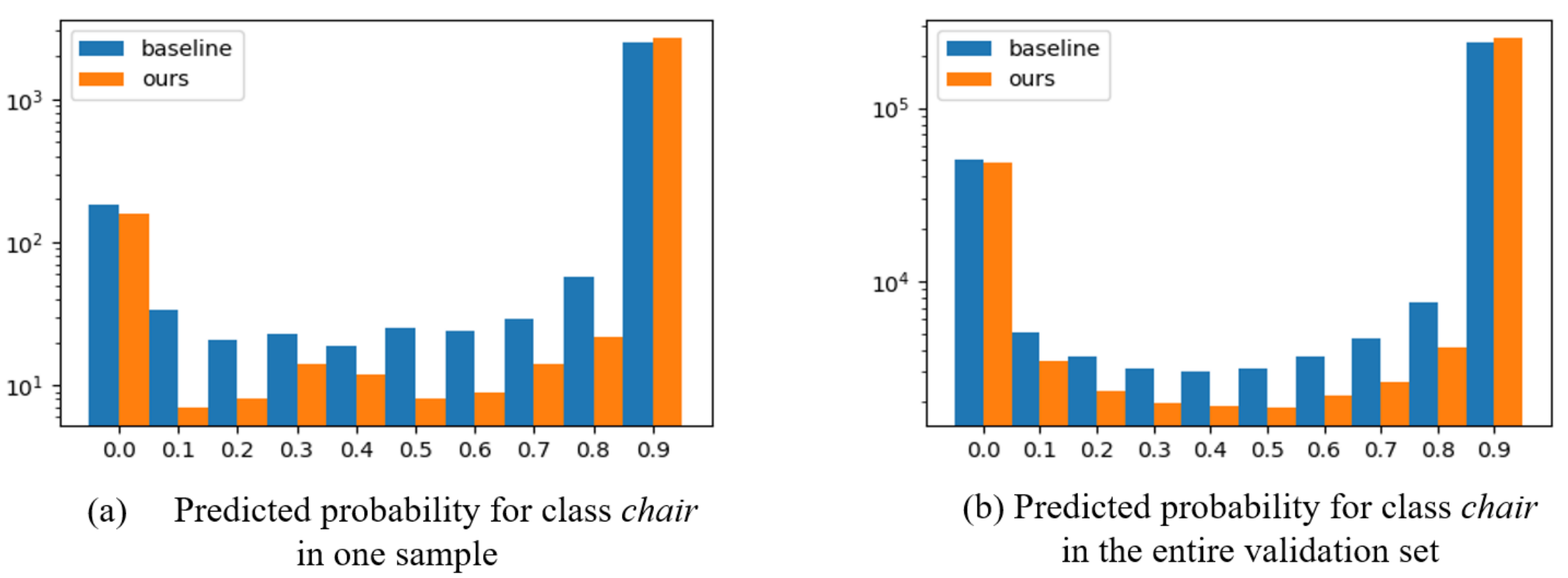}
	\caption{Histogram of predicted probabilities on NYUCAD dataset in log scale.
	}
	\label{fig:hists}
\end{figure}

\subsection{Numerical Analysis of Class Consistency}
Consider an instance that is partially predicted as \textit{chair} and partially as \textit{sofa}. It means that the confidence or the predicted probability is not high enough. This is shown in Figure~\ref{fig:hists} where we visualize the predicted probability, which is obtained by applying a softmax function on the network output, for \textit{chair} for one sample and the entire validation set. The baseline method would produce relatively more predictions of probabilities around 0.5, showing its uncertainty which would lead to inconsistent completion results. With the aid of our two-stage paradigm and classwise entropy loss function, the predicted probabilities are pushed to the extremes, i.e., 0,1. This validates that our methods can indeed produce consistent results no matter it is correct or wrong predictions.

\subsection{Ablation study}

\subsubsection{The effectiveness of FCM and reusing TSDF features}
We first design three models to validate the effectiveness of the initialization method and resuing TSDF features separately. Due to limited GPU memory, we design a relatively lightweight model A based on the 3D RGB feature completion stage shown in Figure \ref{fig:network}, where we remove the three DDR \cite{li2019rgbd} blocks before generating TF1. We then design model B, where we take model A as the 3D RGB feature completion stage and do not reuse the TSDF features in refined semantic scene completion, i.e., we replicate the TSDF branch in the refined semantic scene completion stage as in the previous stage. Model C takes model A as the 3D RGB feature completion stage and reuses the TSDF features in the refined semantic scene completion stage. Results are shown in Table \ref{tab:table2}. By comparing model  A and model B, we can see that, FCM can boost the performance by 1.4\% on SSC. Moreover, it shows that the dense 3D RGB features can indeed help identify instances that were overwhelmed by dense TSDF features. Looking at model B and model C, our reusing mechanism can not only improve both the preliminary results and the refined results but also reduce memory usage considerably during training.

\begin{table}[t]
	\begin{center}
		\caption{Ablation study of effects of FCM and reusing TSDF features on NYUCAD dataset. Here pre. denotes preliminary results and ref. denotes refined results.}
		\label{tab:table2}
		\begin{tabular}{|c|c|c|c|c|c|} 
			\hline
			Method & FCM & reuse & pre. (SSC) & ref. (SSC) & memory\\
			\hline
			Model A & & &54.5 & - & 5266M\\
			\hline
			Model B & \checkmark &   &      54.3 & 55.9 & 9052M\\
			\hline
			Model C & \checkmark & \checkmark &55.7 & 56.7 & 7871M\\
			\hline
		\end{tabular}
	\end{center}
\end{table}

\begin{table}\tiny
	\begin{center}
		\caption{Ablation study of effects of the weight of our classwise entropy loss in two stages. }
		\label{tab:table4}
		\resizebox{0.4\textwidth}{!}{
			\begin{tabular}{|c|c|c|c|c|c|} 
				\hline
				& & \multicolumn{2}{c|}{NYUCAD(SSC)} & \multicolumn{2}{c|}{NYU(SSC)}\\ \hline
				$\lambda_1$ & $\lambda_2$ & \; pre.\; & ref. & pre. & ref.\\
				\hline
				0   & 0   &   56.7    & 57.5 & 41.0 & 41.5\\
				\hline
				0.5 & 0.5   &  57.9 & \textbf{59.0} & 40.8 & 41.7\\
				\hline
				0.5 & 1     &   57.6     & 58.5 & 41.6 & 41.6\\
				\hline
				1   & 0.5  &  57.9      & 58.4 & 41.5 & 41.8\\
				\hline
				1   & 1   & 58.0 & 58.6 & 41.8 &\textbf{42.3} \\ 
				\hline
		\end{tabular}}
	\end{center}
\end{table}

\subsubsection{The effectiveness and generalizability of our classwise entropy loss to architectures that take RGB and TSDF as inputs}
For better illustration, we first provide results on our model as shown in Table \ref{tab:table4}. We experiment on the model in Figure \ref{fig:network} and conduct ablation study on the choice of hyperparameters $\lambda_1$ and $\lambda_2$. With our loss applied, SSC performance can be boosted on both NYU and NYUCAD datasets. We obtain the best performance using $\lambda_1=0.5, \lambda_2=0.5$ on NYUCAD  and $\lambda_1=1, \lambda_2=1$ on NYU.

To examine whether our CEL can be applied to other networks that take RGB and TSDF as input, we apply it to SketchNet \cite{chen20203d}. Results are shown in Table \ref{tab:table5}. Since these methods only produce one SSC result, we use $\lambda$ to indicate the weight of our proposed loss. Also, we refer to a network the same as the 3D RGB feature completion stage in Figure \ref{fig:network} but trained alone as the baseline. Our loss can boost the performance of SketchNet \cite{chen20203d} by 1.1\%  and our baseline by 1.6\%
on NYUCAD dataset. Results in Table \ref{tab:table4} and Table \ref{tab:table5} validate the effectiveness of the proposed loss on our model and other models if the modality distributions in 3D space are different. i.e., RGB-TSDF or RGB-Sketch. Nevertheless, $\lambda$ should be carefully selected for better performance.

\begin{table}[t]\tiny
	\begin{center}
		
		\caption{Ablation study of effects of our proposed classwise entropy loss function on different network architectures.}
		\label{tab:table5}
		\resizebox{0.5\textwidth}{!}{
			\begin{tabular}{|c|c|c|c|} 
				\hline
				Method& $ \lambda=0 $ & $\lambda=0.5$  & $ \lambda=1 $ \\
				\hline
				SketchNet \cite{chen20203d} & 55.2 & 56.1 & 56.3\\
				\hline
				Baseline & 56.1 & 57.5 & 57.7  \\ 
				\hline
		\end{tabular}}
		
	\end{center}
\end{table}

\begin{figure}
	\centering
	\includegraphics[width=0.8\textwidth]{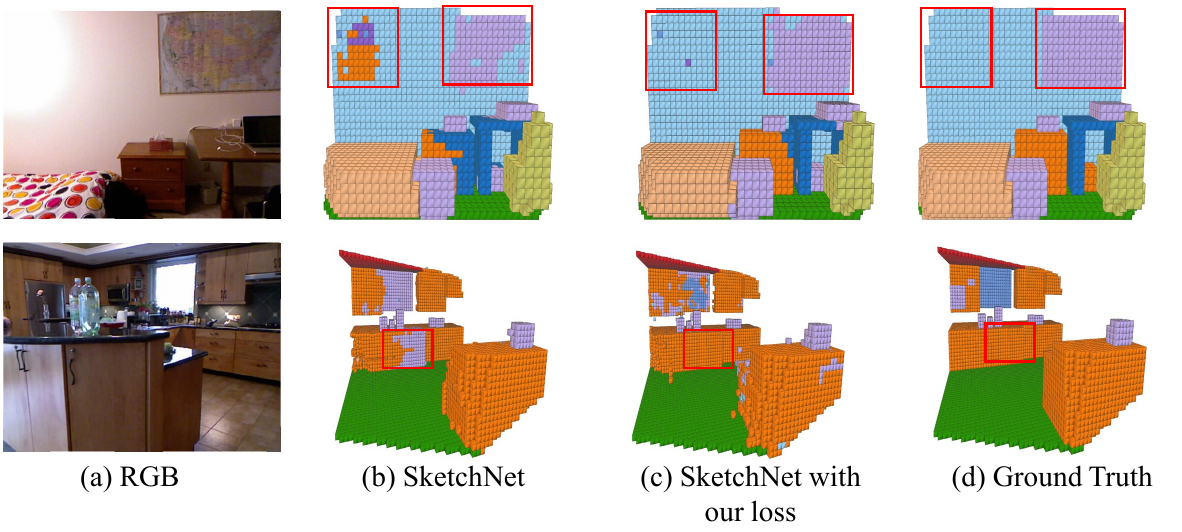}
	\caption{Visualization results comparing applying our loss to SketchNet \cite{chen20203d} or not. With our classwise entropy loss, SketchNet \cite{chen20203d} can produce more consistent results.}
	\label{fig:ablation_loss}
\end{figure}

We provide visualization results on NYUCAD dataset, comparing SketchNet \cite{chen20203d} trained with our proposed loss or not in Figure \ref{fig:ablation_loss}. For the first row, our loss can help produce more consistent results on visible walls. As for the second row, our loss helps achieve consistent completion results shown in red boxes. Unseen areas can be predicted extremely differently from their visible counterparts, yet our loss can help alleviate such inconsistency.

\section{Conclusion}
In this paper, we identify that different distributions between sprase RGB features and dense TSDF features in 3D space can lead to inconsistent predictions. To alleviate this inconsistency, we propose a two-stage network with FCM, which transforms 3D RGB features from sparse to dense. Besides, a novel classwise entropy loss function is introduced to punish inconsistency. Experiments demonstrate the effectiveness of our method compared with our baseline methods and other state-of-the-art methods on public benchmarks. Mainly, our methods can produce consistent and reasonable results.\\

\noindent\textbf{Acknowledgment}
This work was partially supported by Shenzhen Science and Technology Program (JCYJ20220818103006012, ZDSYS20211021111415025), Shenzhen Institute of Artificial Intelligence and Robotics for Society, and the Research Foundation of Shenzhen Polytechnic University (6023312007K).

\bibliographystyle{splncs04}
\bibliography{prcv23}

\end{document}